%% file: paper.tex
\documentclass{article}
\usepackage{spconf,amsmath,graphicx,subfig,textcomp}
\usepackage{times}
\usepackage{epsfig}
\usepackage{amssymb}
\usepackage{csquotes}
\usepackage{caption}
\usepackage{textcomp}
\usepackage{booktabs}

\title{Learning eating environments through scene clustering}
\name{$\text{S. K. Yarlagadda}^{1}$, $\text{Sriram Baireddy}^1$, $\text{David G\"{u}era}^{1}$, \textit{$\text{Carol J. Boushey}^2$, $\text{Deborah A. Kerr}^3$, $\text{Fengqing Zhu}^1$} }
\address{$\text{School of Electrical and Computer Engineering, Purdue University, West Lafayette, IN, USA}^1$ \\ $\text{University of Hawaii Cancer Center, Honolulu, HI, USA}^2$ \\ $\text{School of Public Health, Curtin University, Perth, Western Australia}^3$ }

\begin{document}
%\ninept
%
\maketitle
\begin{abstract}
It is well known that dietary habits have a significant influence on health. While many studies have been conducted to understand this relationship, little is known about the relationship between eating environments and health. Yet researchers and health agencies around the world have recognized the eating environment as a promising context for improving diet and health. In this paper, we propose an image clustering method to automatically extract the eating environments from eating occasion images captured during a community dwelling dietary study. Specifically, we are interested in learning how many different environments an individual consumes food in. Our method clusters images by extracting features at both global and local scales using a deep neural network. The variation in the number of clusters and images captured by different individual makes this a very challenging problem. Experimental results show that our method performs significantly better compared to several existing clustering approaches. 
\end{abstract}
\input{intro.tex}
\input{dataset_related_work.tex}
\input{method.tex}
\input{experiments.tex}
\input{conclusion.tex}
\bibliographystyle{IEEEbib}
\bibliography{refs}

\end{document}

%% file: intro.tex
\section{Introduction}
\label{sec:intro}

In 2016, \$7.5 trillion was spent on healthcare worldwide, which is approximately 10\% of the world GDP~\cite{world2018public}. While there are many factors that influence health, dietary habits have a significant impact~\cite{Mesas2012, Nordstrom2013}.  To understand the complex relationship between dietary habits and health, nutrition practitioners and researchers often conduct dietary studies to subjectively assess dietary intake of children and adults. Participants in these studies are asked to report the foods and drinks they consumed on a daily basis for a period of time. This data is then analyzed by researchers to understand the impact of certain dietary behaviors on health. For example, studies have shown that frequent consumption of fast food~\cite{fast-food}, skipping breakfast~\cite{breakfast}, and absence of home food~\cite{homefood_absence} contribute to the increasing risk of obesity and overweight.

While many studies have been conducted to understand how diet affects health~\cite{food_computing}, fewer work has been done to study the relationship between eating environments and health. However, researchers~\cite{environment1, environment2} and organizations such as the World Heath Organization and International Obesity Task Force have recognized the vital role of eating environments on health and diet. Studies have shown that some factors of the environment such as screen viewing during meals, family mealtime~\cite{dinner}, and meal preparation time~\cite{mealtime} influence health. For instance, family mealtime is shown to positively affect nutrient intake and meal preparation time is inversely related to Body Mass Index (BMI). While such patterns have been found, the relationship between eating environments and health is still poorly understood, partly due to the lack of valid, reliable measures of environmental factors. In this paper, we focus on understanding eating environments of an individual using digital images captured during eating occasions. In particular, we are interested in learning how many different environments an individual consumes food in.

\begin{figure}%
    \centering
    \subfloat[Scene 1]{{\includegraphics[width=6cm,keepaspectratio]{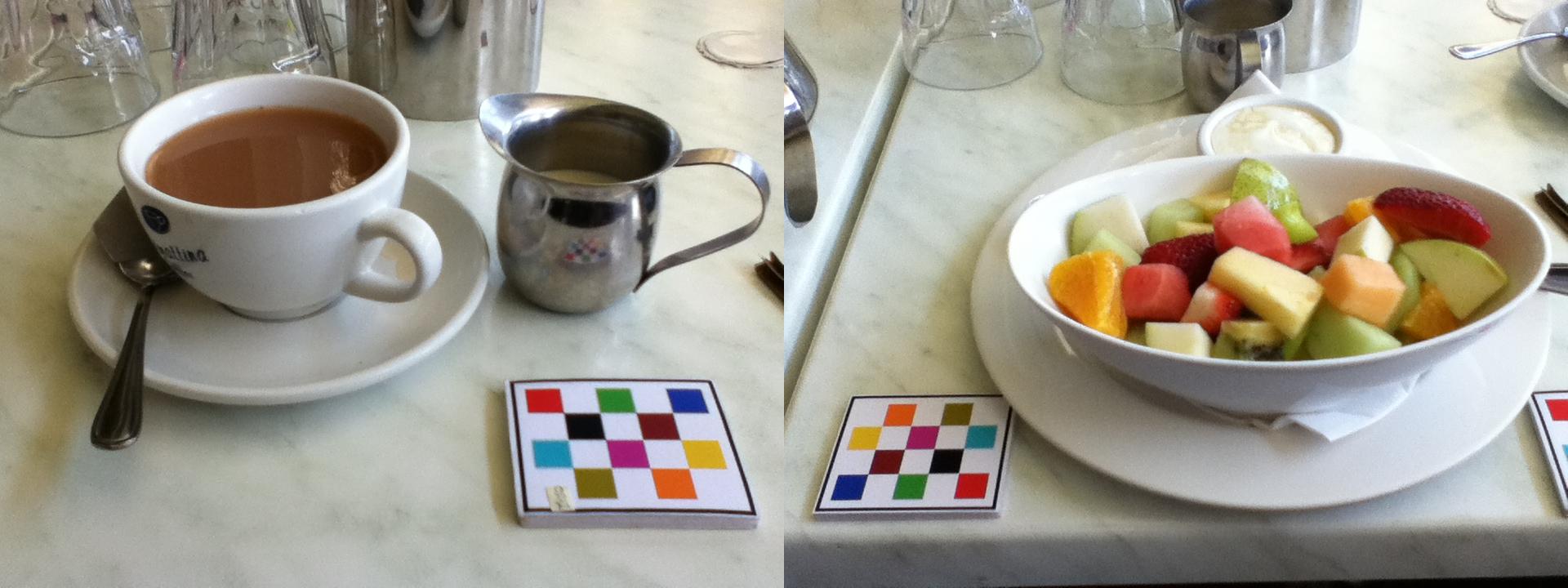} }}%
    \qquad
    \subfloat[Scene 2]{{\includegraphics[width=6cm]{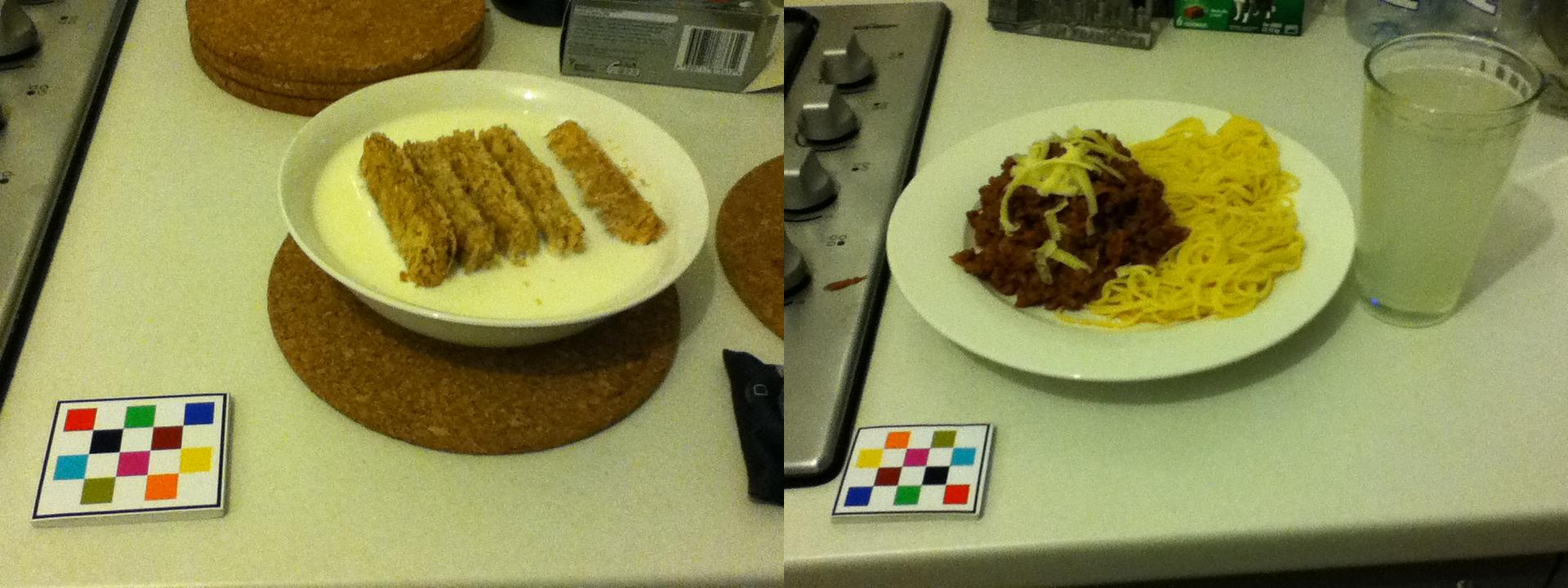} }}%
    \caption{ Here are images of two different eating environments, captured by a single participant. The colored checkerboard in all the images is the FM.}%
    \label{fig:example}%
\end{figure}

Dietary Recall, 24-hr recall, and Food Frequency Questionnaire (FFQ)~\cite{traditional_diteary_assessment_methods} are well-known dietary assessment methods used in most dietary studies. These methods require participants to manually enter details of their diet information through a web interface, or an in-person or phone interview. These methods are known to be time-consuming and prone to errors because participants may not recall all foods and beverages they consumed or cannot accurately estimate the food portion~\cite{traditional_diteary_assessment_methods}. To overcome these limitations, researchers have leveraged advances in mobile technology to develop image-based dietary assessment methods to record daily food intake. Some examples of image based dietary assessment tools are TADA\textsuperscript{\texttrademark}~\cite{Zhu2010} and Food-Log~\cite{food_log}.% FoodCam~\cite{joutou2009}, DietCam~\cite{kong2012}, and Im2Calories~\cite{Meyers_2015_ICCV}.
In these approaches, participants record their diet by capturing images of foods and beverages consumed using mobile cameras. These images can then be analyzed by trained analysts~\cite{Howes2017} or using image analysis methods~\cite{zhu-jbhi2015,fang-nu2019,wang-mta2017} to extract nutrient information. In this paper, we leverage eating occasion images captured using the TADA\textsuperscript{\texttrademark} system to cluster eating scenes based on their visual appearance to help understand the relationship between a person's eating environment and dietary quality. 

\begin{figure}[t!]
    \centering
    \includegraphics[scale = 0.55]{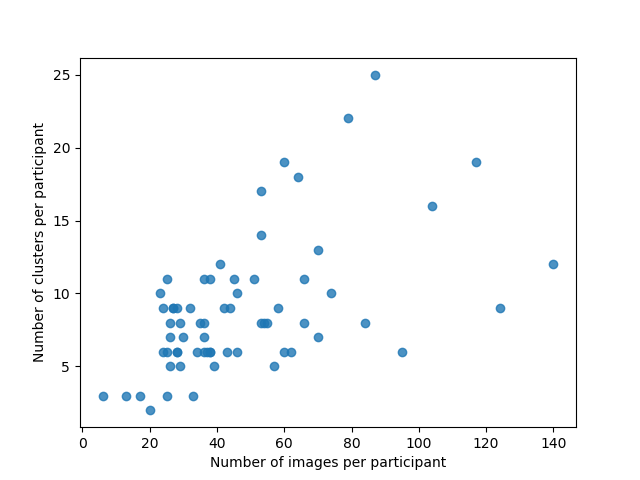}
    \caption{Scatter plot of the number of images captured per participant versus the number of eating scene clusters per participant.}
    \label{fig:scatter_plot}
\end{figure}

%% file: dataset_related_work.tex
\section{Dataset and Related Work}
\begin{figure*}
    \centering
    \includegraphics[scale = 0.4]{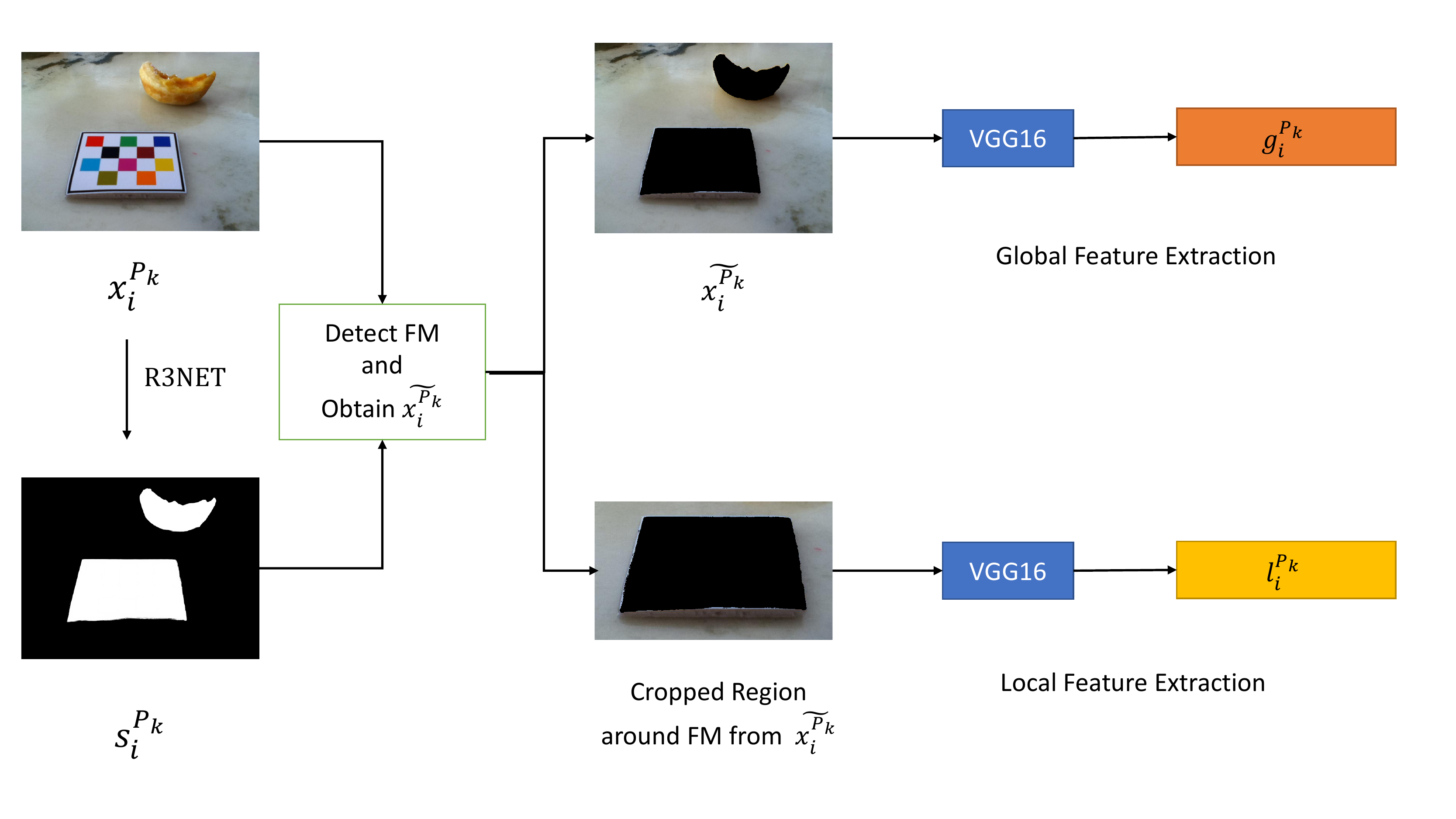}
    \caption{Overview of feature extraction}
    \label{fig:myftr_extr}
\end{figure*}

We used a dataset $\mathcal{D}$ that consists of $3137$ images from 66 participants collected in a community dwelling dietary study~\cite{dataset}. In this study, participants were asked to take pictures of their foods and to place a colored checkerboard with known dimensions, called the Fiducial Marker (FM), in the scene. An example of two pairs of images belonging to two different eating environments are shown in Fig. \ref{fig:example}. The FM serves two purposes: to aid in color correction, and more importantly, to provide a reference scale for food portion estimation, so the nutrient content of the foods can be computed. One of the challenges of this dataset is the large variance in the number of images and the number of eating environments for different participant, as shown in Fig. \ref{fig:scatter_plot}.

Our goal is to cluster food images of a participant based on eating environments. Clustering is not a new concept. Several classical~\cite{survey_classical} and deep learning-based~\cite{survey_dl} methods have been proposed. Classical clustering approaches are applied on relevant features extracted from the data~\cite{clustering_imgs}. However, the problem of clustering food images based on the eating environments captured has not been studied before, so the relevant features are not defined. On the other hand, deep learning-based methods, given sufficient data, are capable of simultaneously learning to extract relevant features and cluster images~\cite{survey_dl}. A common assumption of such approaches is that the number of clusters is known \textit{a priori}~\cite{survey_dl}. However, in our case, we do not know \textit{a priori} the number of eating environments for each participant. In addition, the number of images collected by each participant is not sufficient to apply deep learning-based methods directly, as they usually require hundreds of images per cluster for training a good model.
As both classical and deep learning-based approaches have their  shortcomings for clustering eating environments from images, new techniques need to be developed to solve this problem.

%% file: method.tex
\section{Method}\label{method}

In this section, we describe the details of the proposed method. First, we introduce the notation used throughout the paper. $P_k$ denotes the $k^{\text{th}}$ participant in the dataset. The $i^{\text{th}}$ image captured by participant $P_k$ is denoted by $x^{P_k}_i$. %The ground truth cluster label for $x^{P_k}_i$ is denoted by $y^{P_k}_i$ . 
Our goal is to cluster food images based on eating environment. We do this by extracting features at the local and global level from relevant pixels of $x^{P_k}_i$ and then applying a clustering method on said features.

\subsection{Global Feature Extraction}\label{global}
The image $x^{P_k}_i$ contains many salient objects such as the food, drinks, FM, and silverware. However, pixels belonging to the salient objects do not contain any credible information regarding the eating environment because a person can eat different food with different plates in the same eating environment. Pixels belonging to the FM are also not relevant because the FM is common to all food images, irrespective of the eating environment. Instead, we are interested in pixels belonging to the non-salient regions. To extract these relevant pixels in $x^{P_k}_i$, we first extract its binary saliency map denoted by $s^{P_k}_i$ using a state-of-the-art saliency estimator R3NET \cite{r3net}. The salient pixels of $x^{P_k}_i$ have a value of 1 in $s^{P_k}_i$ and the rest have a value of 0. The relevant pixels of $x^{P_k}_i$ are captured by $\Tilde{x}^{P_k}_i$, and is defined as 
\begin{equation} \label{eq:eq1}
\Tilde{x}^{P_k}_i = (1 - s^{P_k}_i)*x^{P_k}_i
\end{equation}
Features are extracted from $\Tilde{x}^{P_k}_i$ using a Convolutional Neural Network (CNN) VGG16 \cite{vgg16} pre-trained on the ImageNet dataset~\cite{imagenet_cvpr09}. We use a pre-trained VGG16 for feature extraction because these features are robust to artifacts such as noise, lighting changes, and differing viewpoints. 

The $2^{nd}$ convolutional layer is chosen for feature extraction and the reasoning behind this is explained in Section \ref{hp-tuning}. Global Average Pooling (GAP) is applied to the output of the $2^{nd}$ layer to spatially summarize the information into a 64-dimensional vector. This vector is denoted as $g^{P_k}_i$ and we refer to it as the global feature.

\subsection{Local Feature Extraction}
Since the FM is always placed in close proximity to the foods, we assume it is on the same surface as the foods. Examples of such surfaces may include desk, dining table, and kitchen counter, to name a few. These surfaces give us a lot of information about the eating environment because it is very likely that a person uses the same surface during an eating event in a particular eating environment. Therefore, we extract features from this surface and they are denoted by $l^{P_k}_i$. 

By identifying the FM in the image, we can extract information about the eating surface. The FM is detected by finding the salient object with the highest number of interest points. An interest point is a pixel that ORB (Oriented FAST and Rotated BRIEF)~\cite{ORB} finds useful for image registration. The FM has a lot of interest points because of the colored checkerboard pattern and it is unlikely for other salient objects such as foods to have as many distinct interest points. The salient objects of the image can be found by performing connected component analysis on $s^{P_k}_i$ and treating each connected component as a salient object. Once we have located the FM, we extract a region around it from $\Tilde{x}^{P_k}_i$. $l^{P_k}_i$ is obtained from this region using the pre-trained VGG16 in the same way as done in \ref{global}. Fig. \ref{fig:myftr_extr} illustrates the the global and local feature extraction process.

\subsection{Feature Fusion and Clustering}
We fuse the local and global features using their distance matrices. The distance matrices $G^{P_k}$ and $L^{P_k}$ are defined as follows 
\begin{equation} \label{eq:eq1}
\begin{split}
L_{(i, j)}^{P_k} = ||l^{P_k}_i - l^{P_k}_j||_2 \\
G_{(i, j)}^{P_k} = ||g^{P_k}_i - g^{P_k}_j||_2
\end{split}
\end{equation}
The fused distance matrix $D^{P_k}$ is defined as follows 
\begin{equation}\label{eq:eq2}
D^{P_k} = \alpha*L^{P_k} + (1 - \alpha)*G^{P_k}
\end{equation}
Here $\alpha \in [0, 1]$ and controls the relative importance of the local and global features. A higher value of $\alpha$ indicates local features are more important and vice versa. In our case , $\alpha = 0.44$. The reason behind this is explained in Section \ref{hp-tuning}.  Affinity Propagation (AP)~\cite{AP} is applied to $D^{P_k}$ to obtain the final clusters. 

%% file: experiments.tex
\section{Experiments}
\begin{figure}
    \centering
    \includegraphics[scale = 0.55]{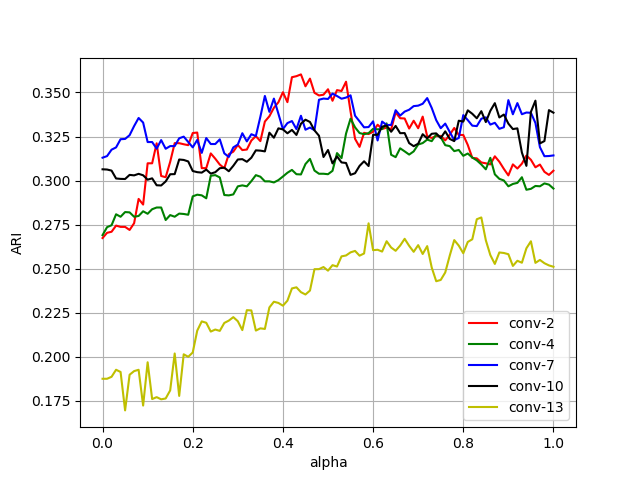}
    \caption{ARI scores for different $\alpha$ and convolutional layer $m$ on $\mathcal{D}_{\text{val}}$}
    \label{fig:ari_plot}
\end{figure}
In this section we describe all the experiments conducted and compare our method to four existing clustering methods, namely: DBSCAN~\cite{dbscan}, MeanShift~\cite{meanshift}, OPTICS~\cite{optics}, and AP~\cite{AP}. We use the Adjusted Rand Index (ARI) and Normalized Mutual Information (NMI) to measure the accuracy of our clusters. ARI ranges from $-1.0$ to $1.0$ while NMI ranges from $0$ to $1.0$. Higher values indicate better clustering for both ARI and NMI. Accuracy over a dataset is reported as the average accuracy among all participants.

\subsection{Hyperparameter Tuning}\label{hp-tuning}
Our method has two hyperparameters: the weighting factor for feature fusion $\alpha$ and the convolutional layer of VGG16 for feature extraction $m$. Our dataset $\mathcal{D}$ is split into $\mathcal{D_{\text{val}}}$ and $\mathcal{D_{\text{test}}}$. $\mathcal{D_{\text{val}}}$ consists of images from 10 participants and $\mathcal{D_{\text{test}}}$ contain images from 56 participants. Our method is evaluated on $\mathcal{D_{\text{val}}}$  by varying $\alpha$ and $m$. $\alpha$ ranges from 0 to 1 in steps of $0.01$. To find the optimal $m$ for our dataset, we extract features from convolutional layers preceding a max-pooling layer. There are five such layers in VGG16. ARI is used to select the optimal values. Fig.~\ref{fig:ari_plot} shows the ARI scores for different values of $\alpha$ and convolutional layer $m$, and the optimal value for $\alpha$ is 0.44 and for $m$ is $2$. 

From equation \ref{eq:eq2}, we can infer that as the value of $\alpha$ increases, more weight is given to local features and vice-versa. An optimal value of $0.44$ for $\alpha$ suggests that our method performs best when approximately  equal weight is given to both local and global features. This shows that using only one of these features is less optimal. In Fig.~\ref{fig:ari_plot}, we can see that the performance of our method degrades once the chosen feature extraction layer gets very deep. Later layers, like conv-13, extract abstract features and completely lose information about edges, colors, and textures. We suspect this loss of information is the reason for decrease in performance.

\subsection{Testing}
We evaluated the performance of our method on $\mathcal{D_{\text{test}}}$ after selecting the optimal values for $\alpha$ and $m$ using the validation set $\mathcal{D_{\text{val}}}$. We choose four well-known clustering methods for comparison, namely DBSCAN~\cite{dbscan}, MeanShift~\cite{meanshift}, OPTICS~\cite{optics}, and AP~\cite{AP} using the eating scene image as the input. ARI and NMI for the five methods are reported in Table \ref{tb:auc_f}, where our method achieves the best performance. It is worth noting that although we use AP~\cite{AP} for clustering after feature fusion, our method performs significantly better than AP~\cite{AP} when meaningful features are not known. This indicates that our feature extraction strategy is highly relevant and very important to our problem. 

\begin{center}
\captionof{table}{ARI and NMI scores for methods tested on  $\mathcal{D_{\text{test}}}$. The best results are reported in bold.}
\begin{tabular}{ccc}
	\toprule
	Methods & ARI & NMI	 \\
	\midrule
	Ours & \textbf{0.39} & \textbf{0.68} \\
	DBSCAN~\cite{dbscan} & 0.24 & 0.47 \\
	MeanShift~\cite{meanshift} & 0.12 & 0.35 \\
    OPTICS~\cite{optics} & 0.08 & 0.39 \\
	AP~\cite{AP} & 0.2 & 0.49 \\
    \bottomrule
\end{tabular}
\label{tb:auc_f}
\end{center}

%% file: conclusion.tex
\section{Conclusion}
We proposed a method to cluster food images based on their eating environment. Our method extracts features from a pre-trained CNN at multiple levels. These features are fused using their distance matrices and a clustering algorithm is applied after feature fusion. Our method is evaluated on a dataset containing 3137 eating scene images collected from a dietary study with a total of 585 clusters. We compared our method to state-of-the-art clustering methods and showed improved performance.